\documentclass{article}
\usepackage{spconf,amsmath,graphicx,hyperref}
\usepackage{multirow}
\usepackage{booktabs}
\usepackage{tipa}

\title{Child-Adapted Structured Phonological Representations for Interpretable Speech Sound Analysis}
\name{
  Abner Hernandez$^{1}$,
  Tom\'as Arias-Vergara$^{1,2}$,
  Andreas Maier$^{1}$,
  Paula~A. Perez-Toro$^{1,2,3}$
}

\address{
  $^{1}$ Pattern Recognition Lab, Friedrich-Alexander-Universität Erlangen-Nürnberg, Germany \\
  $^{2}$ GITA Lab. Facultad de Ingeniería. Universidad de Antioquia UdeA, Medellín, Colombia\\
  $^{3}$ Chair for AI in Healthcare and Medicine, Technical University of Munich, Munich, Germany
}
\begin{document}
%
\maketitle
\begin{abstract}
Structured phonological representations provide an interpretable alternative to generic speech embeddings, but existing models are largely trained on adult speech. We adapt PhonoQ-2.0 to child speech using CHILDES-Aligned data and compare three alignment-supervision conditions (Adult, Adult+Child, and Child-only) across two initialization strategies (Adult PhonoQ and scratch). Generalization is evaluated against manual child-speech annotations. On 1,352 consonant targets from 58 typically developing children, child-speech adaptation improves voicing recognition across all supervision conditions, from 0.922 macro-F1 for Adult PhonoQ to 0.972--0.987 after adaptation. Manner is more sensitive to alignment supervision: Adult+Child MFA reaches 0.804 and 0.796, compared to approximately 0.70 under Adult MFA supervision. Place remains comparatively strong across systems (0.871--0.902), although per-class performance varies substantially. The velar-fronting contrast is preserved across all seven model variants. Longitudinal UltraPhonix analysis further reveals speaker-specific velar and post-alveolar changes that are largely preserved across models and broadly consistent with reported clinical progress.
\end{abstract}
\section{Introduction}
\label{sec:intro}

Self-supervised speech models provide strong general-purpose representations for phonetic and phonological modelling~\cite{Baevski2020,Hsu2021,Chen2022}. However, most pretrained models and downstream systems are developed primarily from adult speech. Child speech differs from adult speech in vocal-tract characteristics, acoustic variability, and ongoing phonological development, making direct transfer challenging. These differences are particularly important when the goal is not only transcription, but analysis of how individual speech sounds and phonological contrasts are realized.

Recent work has therefore investigated the adaptation of self-supervised models to child speech. Studies of wav2vec~2.0, HuBERT, and WavLM show that child ASR remains sensitive to domain mismatch and benefits from child-specific adaptation~\cite{Fan2024,BlockMedin2024}. Beyond transcription, phoneme-recognition errors have also been used to study developmental pronunciation patterns~\cite{Horii2025}, while learned speech representations have been applied to pronunciation assessment and speech sound disorder analysis~\cite{Getman2022,Ng2022,Sung2025}. These approaches demonstrate the potential of learned representations for child speech, but typically rely on discrete phoneme recognition, transcription, global assessment scores, or disorder-level classification.

Structured phonological representations offer a complementary approach by describing speech through properties such as manner and place of articulation, voicing, and vowel quality. Recent work further suggests that these properties are intrinsically structured within self-supervised speech representations: Choi et al.~\cite{Choi2026} identified linear directions corresponding to phonological features across languages and showed that their magnitude reflects continuous differences in acoustic realization. PhonoQ and PhonoQ-2.0 make this information explicit by directly predicting frame-level phonological features from speech~\cite{arias2022analysis,arias2023measuring,Hernandez2026PhonoQ}. However, the English PhonoQ-2.0 model was trained entirely on adult speech, leaving open how well these structured representations transfer to children.

In this work, we adapt PhonoQ-2.0 to child speech using CHILDES-Aligned~\cite{Zheng2026CHILDESAligned}. We study the source of phonetic supervision using Adult, Adult+Child, and Child-only forced alignments, and compare Adult PhonoQ initialization with scratch initialization of the PhonoQ-specific components. Because alignment-derived labels do not provide a common independent ground truth across supervision conditions, we evaluate generalization on manually annotated child speech from UltraSuite~\cite{Eshky2018}. We then examine whether the resulting place-of-articulation posteriors capture velar fronting and longitudinal velar and post-alveolar fronting in UltraPhonix intervention data~\cite{Cleland2019,Cleland2021}. Together, these experiments evaluate whether child-specific supervision improves structured phonological recognition while preserving interpretable speech-sound contrasts.

\section{Methods}
\label{sec:methods}

\subsection{PhonoQ-2.0 Child Adaptation}
\label{sec:phonoq_adaptation}

PhonoQ-2.0 combines a wav2vec~2.0 XLSR encoder with a two-layer Conformer and four structured prediction heads: manner (9 classes), vowel height/backness (6 outputs), place (5 classes), and voicing (2), yielding a 22-D frame-level representation. The original English model was trained on approximately 54~h of adult CommonPhone, TIMIT, and LibriSpeech speech. Child adaptation retains the same architecture and inventory and uses a speaker-disjoint CHILDES-Aligned subset of 44.77~h/36,577 training, 2.77~h/2,277 development, and 2.72~h/2,360 test utterances~\cite{Zheng2026CHILDESAligned}.

\subsection{Alignment Supervision and Initialization}
\label{sec:alignment}

\begin{figure}[t]
\centering
\includegraphics[width=0.90\columnwidth]{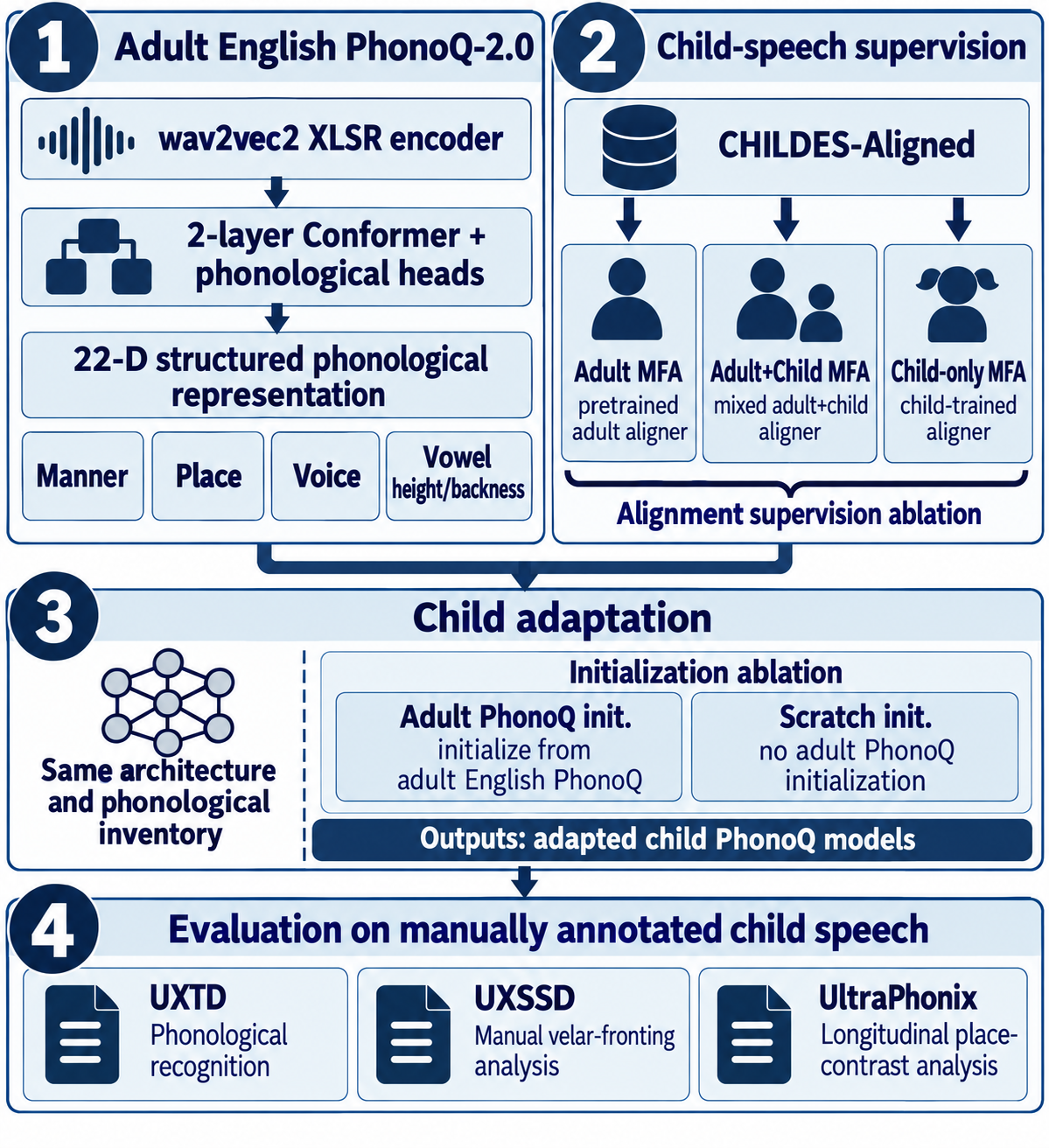}
\caption{Overview of PhonoQ-2.0 child-speech adaptation and downstream evaluation.}
\label{fig:pipeline}
\end{figure}

Figure~\ref{fig:pipeline} summarizes the $3\times2$ adaptation design: Adult, Adult+Child, and Child-only MFA supervision crossed with Adult- and Scratch-init. The three MFA systems use pretrained adult English, approximately 143~h mixed adult--child, and approximately 90~h child speech, respectively, and provide frame-level labels for the same CHILDES-Aligned subset. No UltraSuite speech is used for adaptation or MFA training. Adult-init starts from Adult PhonoQ-2.0; Scratch-init retains the pretrained XLSR encoder but randomly initializes the PhonoQ-specific components.

Because each MFA condition produces its own labels, held-out CHILDES-Aligned scores assess only how consistently each supervision source is learned; cross-condition generalization is evaluated using manual annotations.

MFA intervals are mapped to the 22-D PhonoQ inventory and aligned to XLSR frames by centre-frame mapping. Training uses class-weighted cross-entropy for manner, vowel height, vowel backness, place, and voicing, with label smoothing of 0.05. The lower 20 of 24 XLSR layers are frozen; models are trained for up to 30 epochs with learning rates of $10^{-5}$/$10^{-3}$ for XLSR/PhonoQ components, weight decay $0.01$, effective batch size 16, mixed precision, gradient clipping at 0.5, and model selection by development macro-F1 with early-stopping patience 10.

\subsection{Independent Phonological Recognition on UXTD}
\label{sec:uxtd_methods}

Generalization is evaluated on 1,352 manually annotated consonant landmarks from 58 UXTD speakers, provided by speech and language therapists (SLTs)~\cite{Eshky2018}. Each landmark marks the temporal location of a target consonant. Manner, place, and voicing are evaluated for all targets; vowel features are not evaluated. Posteriors are averaged within $\pm25$~ms of each landmark, and macro-F1 is computed over reference classes with non-zero support. Uncertainty is estimated using 20,000 speaker-cluster bootstrap resamples.


\subsection{Manual Velar-Fronting Analysis}
\label{sec:uxssd_vf}

Using the released SLT annotations, we compare manually localized /k/ productions from four UXSSD children with velar fronting and 10 age-matched UXTD controls. Place posteriors are averaged within $\pm50$~ms and scored as

\begin{equation}
s_{\mathrm{VF}}
=
\bar{P}(\mathrm{alveolar})
-
\bar{P}(\mathrm{velar}),
\label{eq:vf_score}
\end{equation}

where positive values indicate greater alveolar evidence. Speakers are summarized by the median token score, and groups are compared using the between-group median difference, 95\% speaker-bootstrap CI, AUC, and an exact two-sided permutation test.

\newcommand{\se}[1]{\mbox{\footnotesize $\pm$\,#1}}

\begin{table*}[t]
\centering
\caption{Per-class F1 on manually annotated UXTD consonant targets. Values are point estimates with bootstrap standard errors from 20,000 speaker-cluster resamples. A-init and S-init denote Adult- and Scratch-init, respectively.}
\label{tab:uxtd_per_class}
\small
\setlength{\tabcolsep}{3.2pt}
\begin{tabular*}{\textwidth}{@{\extracolsep{\fill}}llccccccc@{}}
\toprule
& & \multirow{2}{*}{Adult}
& \multicolumn{2}{c}{Adult MFA}
& \multicolumn{2}{c}{Adult+Child MFA}
& \multicolumn{2}{c}{Child-only MFA} \\
\cmidrule(lr){4-5}
\cmidrule(lr){6-7}
\cmidrule(lr){8-9}
Group & Feature
& PhonoQ
& A-init & S-init
& A-init & S-init
& A-init & S-init \\
\midrule

\multirow{7}{*}{Manner}
& Stop
& 0.72\,\se{0.02}
& 0.80\,\se{0.02}
& 0.79\,\se{0.02}
& \textbf{0.84}\,\se{0.02}
& 0.81\,\se{0.02}
& 0.80\,\se{0.02}
& 0.82\,\se{0.02} \\

& Nasal
& 0.87\,\se{0.06}
& 0.84\,\se{0.08}
& 0.81\,\se{0.10}
& 0.85\,\se{0.06}
& \textbf{0.90}\,\se{0.05}
& 0.83\,\se{0.06}
& 0.84\,\se{0.06} \\

& Rhotic
& 0.68\,\se{0.03}
& 0.72\,\se{0.03}
& 0.68\,\se{0.02}
& 0.73\,\se{0.02}
& 0.72\,\se{0.02}
& \textbf{0.74}\,\se{0.02}
& 0.70\,\se{0.02} \\

& Fricative
& 0.84\,\se{0.02}
& 0.90\,\se{0.02}
& 0.93\,\se{0.01}
& 0.97\,\se{0.01}
& \textbf{0.97}\,\se{0.01}
& 0.97\,\se{0.01}
& 0.96\,\se{0.01} \\

& Affricate
& 0.86\,\se{0.03}
& 0.89\,\se{0.03}
& 0.93\,\se{0.02}
& 0.94\,\se{0.02}
& 0.94\,\se{0.02}
& \textbf{0.95}\,\se{0.02}
& 0.81\,\se{0.04} \\

& Approximant
& 0.49\,\se{0.12}
& 0.33\,\se{0.11}
& 0.15\,\se{0.09}
& 0.57\,\se{0.12}
& 0.44\,\se{0.10}
& 0.33\,\se{0.11}
& \textbf{0.61}\,\se{0.05} \\

& Lateral
& 0.49\,\se{0.07}
& 0.40\,\se{0.07}
& 0.63\,\se{0.06}
& 0.73\,\se{0.06}
& \textbf{0.80}\,\se{0.04}
& 0.73\,\se{0.05}
& 0.67\,\se{0.05} \\

\midrule
\multirow{4}{*}{Place}
& Labial
& 0.72\,\se{0.03}
& 0.74\,\se{0.03}
& 0.76\,\se{0.03}
& 0.74\,\se{0.03}
& \textbf{0.77}\,\se{0.03}
& 0.73\,\se{0.03}
& \textbf{0.77}\,\se{0.03} \\

& Alveolar
& 0.91\,\se{0.01}
& 0.91\,\se{0.01}
& \textbf{0.93}\,\se{0.01}
& 0.91\,\se{0.01}
& 0.92\,\se{0.01}
& 0.92\,\se{0.01}
& 0.91\,\se{0.01} \\

& Velar
& 0.96\,\se{0.01}
& \textbf{0.98}\,\se{0.01}
& 0.98\,\se{0.01}
& 0.97\,\se{0.01}
& 0.94\,\se{0.01}
& 0.98\,\se{0.01}
& 0.96\,\se{0.01} \\

& Postalveolar
& 0.93\,\se{0.02}
& 0.86\,\se{0.03}
& 0.95\,\se{0.02}
& 0.95\,\se{0.02}
& \textbf{0.96}\,\se{0.01}
& 0.96\,\se{0.01}
& 0.88\,\se{0.03} \\

\midrule
\multirow{2}{*}{Voicing}
& Voiceless
& 0.93\,\se{0.01}
& 0.98\,\se{0.01}
& 0.98\,\se{0.00}
& 0.98\,\se{0.00}
& 0.99\,\se{0.00}
& 0.98\,\se{0.00}
& \textbf{0.99}\,\se{0.00} \\

& Voiced
& 0.92\,\se{0.01}
& 0.97\,\se{0.01}
& 0.98\,\se{0.00}
& 0.98\,\se{0.01}
& 0.98\,\se{0.01}
& 0.97\,\se{0.01}
& \textbf{0.99}\,\se{0.00} \\

\bottomrule
\end{tabular*}
\end{table*}

\subsection{Longitudinal UltraPhonix Analysis}
\label{sec:upx_methods}

We analyse SLT-annotated velar- and post-alveolar-fronting targets from six and three UltraPhonix participants, respectively~\cite{Eshky2018,Cleland2019,Cleland2021}. Token-level contrasts are

\begin{equation}
\begin{aligned}
\ell_{\mathrm{VF}}  &= \log P(V)-\log P(A), \\
\ell_{\mathrm{PAF}} &= \log P(PA)-\log P(A),
\end{aligned}
\end{equation}

where $V$, $A$, and $PA$ denote velar, alveolar, and post-alveolar place classes. Target/control phones are /k,g/ versus /t,d/ for velar fronting and /\textipa{S}/ versus /s/ for post-alveolar fronting. For each speaker and session,

\begin{equation}
\Delta =
\operatorname{med}_{\mathrm{target}}(\ell)
-
\operatorname{med}_{\mathrm{control}}(\ell),
\label{eq:upx_delta}
\end{equation}

with larger values indicating stronger target-place separation. $\Delta$ is tracked across baseline, mid-treatment, post-treatment, and maintenance for all seven model variants.

\section{Results}
\label{sec:results}

\subsection{Child-Speech Adaptation}
\label{sec:child_results}

With Adult PhonoQ initialization, held-out macro-F1 increases from 0.685 with Adult MFA supervision to 0.761 with Adult+Child MFA and 0.780 with Child-only MFA. Because each condition is evaluated against its own alignment-derived labels, these scores do not directly compare phonetic alignment accuracy; under Child-only supervision, Adult- and Scratch-init are nearly identical (0.780/0.779).

\subsection{External Phonological Recognition on UXTD}
\label{sec:uxtd_results}

Child-informed supervision shows its clearest effect for manner recognition. Adult PhonoQ and the two Adult-MFA adaptations obtain macro-F1 scores of 0.705, 0.698, and 0.701, whereas Adult+Child MFA reaches 0.804/0.796 and Child-only MFA 0.762/0.775 for Adult-/Scratch-init. The similar pattern under both initialization strategies suggests that manner recognition is more strongly associated with the supervision source than with reuse of the Adult PhonoQ-specific parameters.

Place remains comparatively strong across systems (0.871--0.902) without a monotonic relationship with increasingly child-specific supervision, while voicing increases from 0.922 for Adult PhonoQ to 0.972--0.987 after adaptation. The per-class results in Table~\ref{tab:uxtd_per_class} further show that performance varies within feature groups: velar and postalveolar place are consistently strong, whereas labial is weaker, and rhotic and approximant remain among the more difficult manner classes.

\begin{figure*}[!t]
    \centering
    \includegraphics[width=0.97\textwidth]
    {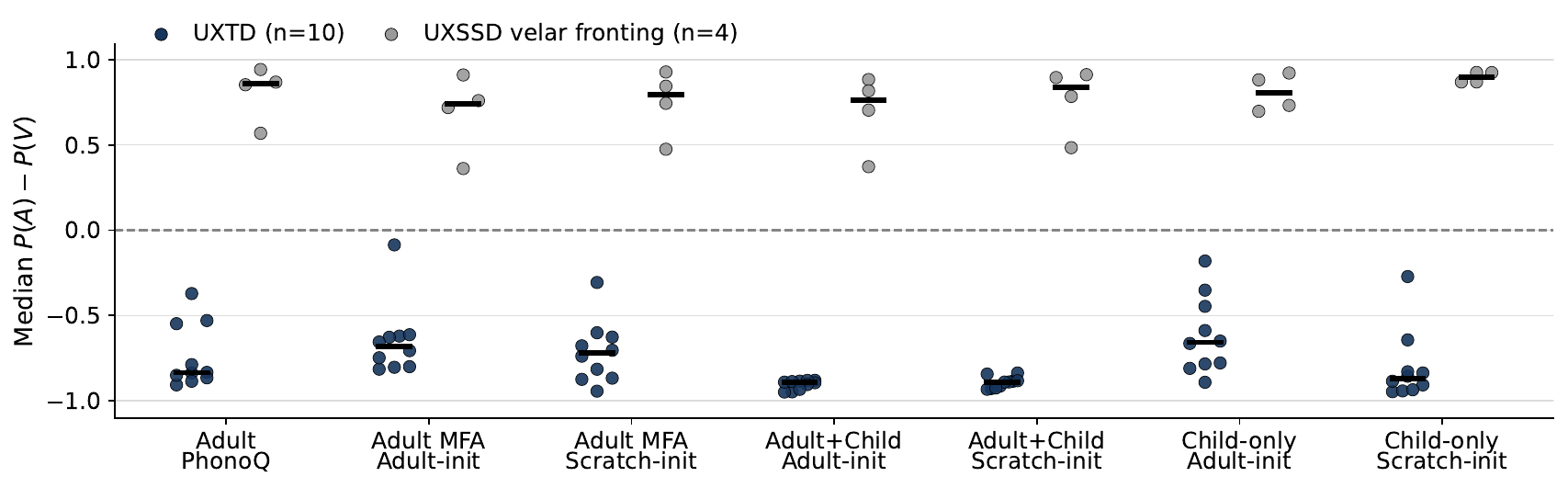}
    \caption{Speaker-level velar-fronting scores for UXSSD children and age-matched UXTD controls across all model variants. Scores are median $\bar{P}(A)-\bar{P}(V)$ contrasts within $\pm50$~ms of the SLT landmark; higher values indicate greater alveolar relative to velar evidence. Horizontal bars show group medians.}
    \label{fig:uxssd_vf}
\end{figure*}


\subsection{Manual Velar-Fronting Analysis}
\label{sec:uxssd_vf_results}

Figure~\ref{fig:uxssd_vf} shows speaker-level velar-fronting scores for the four UXSSD children and 10 age-matched UXTD controls. Across all seven model variants, UXSSD speakers show consistently greater alveolar relative to velar evidence than the typically developing controls.

The between-group median difference ranges from 1.421 to 1.769 across models, with all 95\% speaker-bootstrap confidence intervals excluding zero. In every case, all four UXSSD speaker scores exceed those of all 10 UXTD speakers (AUC$=1.00$), and exact two-sided permutation testing gives $p=0.005$. These results indicate that the velar-fronting contrast is preserved across substantial changes in both alignment supervision and model initialization.

\subsection{Longitudinal UltraPhonix Analysis}
\label{sec:upx_longitudinal_results}

\begin{figure}[t]
    \centering
    \includegraphics[width=\columnwidth]
    {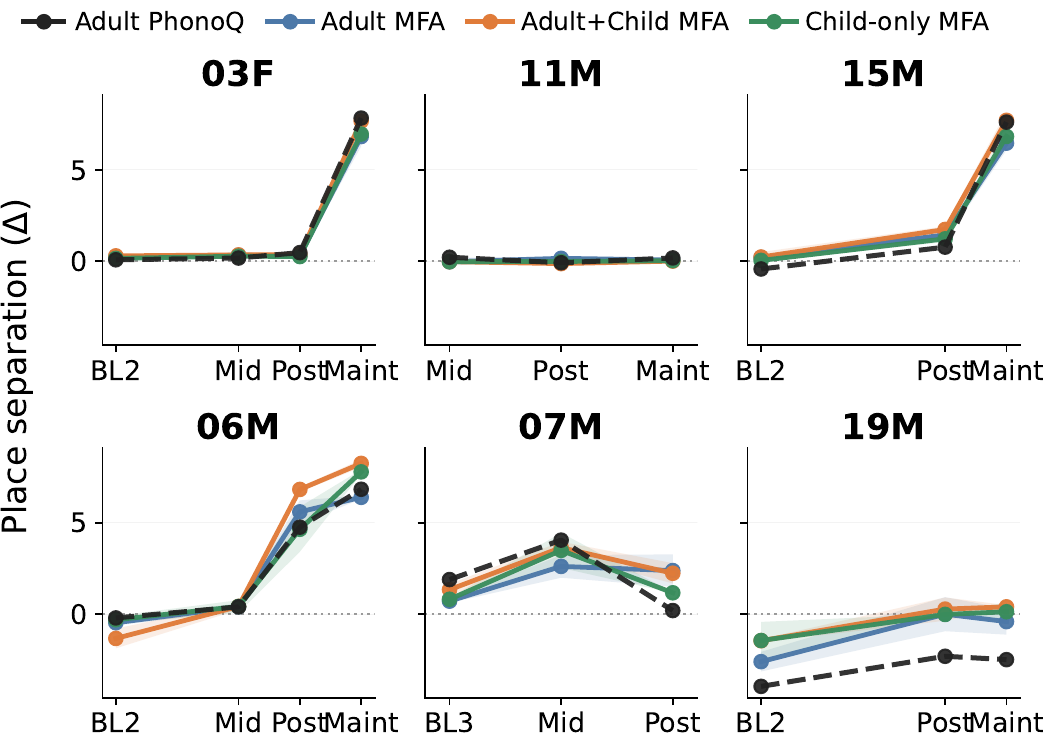}
    \caption{Representative longitudinal trajectories for velar (top) and post-alveolar fronting (bottom). Higher $\Delta$ indicates stronger target-place separation; adapted-model lines average both initializations and shading shows their range.}
    \label{fig:upx_longitudinal}
\end{figure}

Figure~\ref{fig:upx_longitudinal} shows, for space, three of the six velar-fronting participants and all three post-alveolar-fronting participants, with the two initialization variants averaged within each alignment-supervision condition. For velar fronting, 03F and 15M show pronounced increases in target-place separation at later sessions, particularly at maintenance, whereas 11M remains comparatively stable.

For post-alveolar fronting, 06M shows the clearest longitudinal change: baseline mean $\Delta$ ranges from $-1.33$ to $-0.31$ across the adapted conditions, increasing to $4.64$--$6.82$ post-treatment and $6.39$--$8.24$ at maintenance. Speaker 07M also increases from baseline to mid-treatment across all conditions, although the separation decreases again at post-treatment. For 19M, an initially negative contrast approaches zero at post-treatment and remains comparatively small at maintenance. Among the three unplotted velar-fronting participants, 04M also shows a pronounced later increase, whereas 01F and 17M show smaller or less consistent changes.

\section{Discussion and Conclusion}

Taken together, the results suggest that child adaptation affects phonological dimensions differently. Manner is most sensitive to the source of alignment supervision: Adult+Child and Child-only supervision yield higher macro-F1 than Adult-MFA adaptation on manually annotated UXTD speech. The per-class results show that this effect is not uniform, with the clearest gains occurring for several classes including stops, fricatives, and laterals, while rhotics and approximants remain more challenging. Voicing is consistently higher after adaptation to child speech regardless of alignment source, consistent with a broader child-domain effect. Place recognition remains comparatively strong but is also feature dependent, with consistently high velar and postalveolar F1 and lower performance for labial targets.

The longitudinal analysis provides further evidence that these posteriors capture meaningful speaker-specific phonological change. The trajectories broadly follow patterns reported in the UltraPhonix clinical literature: 03F and 15M show strong later improvement in velar contrasts, 11M remains largely stable, and 06M shows a pronounced post-treatment and maintenance increase for post-alveolar fronting. Other participants show weaker or less consistent correspondence, including 07M, whose large clinically reported improvement is less clearly reflected in the PhonoQ trajectory. PhonoQ posteriors should therefore be interpreted as structured acoustic--phonological measures rather than direct estimates of treatment outcome.

Limitations include the absence of a common manual alignment reference for CHILDES, differences in the data used to train the MFA systems, and small clinical cohorts covering only two place-related processes. Future work should extend evaluation to additional phonological features, larger child populations, and a broader range of speech sound processes. Overall, Adult PhonoQ already transfers useful structured information to child speech, while child-speech adaptation yields higher voicing performance and child-informed alignment supervision is particularly beneficial for manner recognition, although gains vary across individual feature classes. Interpretable place contrasts remain robust across model variants.


\section{Compliance with Ethical Standards}

This study is a secondary analysis of previously collected child-speech datasets and involved no new participant recruitment or data collection. The UltraSuite datasets were collected with approval from the NHS Research Ethics Service and with guardian consent for research data sharing~\cite{Eshky2018}. CHILDES-Aligned and all other data were used in accordance with their respective access, data-sharing, and licensing conditions.

\bibliographystyle{IEEEbib}
\bibliography{refs}

@Article{Baevski2020,
  author  = {Baevski, Alexei and Zhou, Yuhao and Mohamed, Abdelrahman and others},
  title   = {wav2vec 2.0: A Framework for Self-Supervised Learning of Speech Representations},
  journal = {Advances in Neural Information Processing Systems},
  volume  = {33},
  pages   = {12449--12460},
  year    = {2020}
}

@Article{Hsu2021,
  author  = {Hsu, Wei-Ning and Bolte, Benjamin and Tsai, Yao-Hung Hubert and others},
  title   = {{HuBERT}: Self-Supervised Speech Representation Learning by Masked Prediction of Hidden Units},
  journal = {IEEE/ACM Transactions on Audio, Speech, and Language Processing},
  volume  = {29},
  pages   = {3451--3460},
  year    = {2021},
  doi     = {10.1109/TASLP.2021.3122291}
}

@Article{Chen2022,
  author  = {Chen, Sanyuan and Wang, Chengyi and Chen, Zhengyang and others},
  title   = {{WavLM}: Large-Scale Self-Supervised Pre-Training for Full Stack Speech Processing},
  journal = {IEEE Journal of Selected Topics in Signal Processing},
  volume  = {16},
  number  = {6},
  pages   = {1505--1518},
  year    = {2022},
  doi     = {10.1109/JSTSP.2022.3188113}
}

@InProceedings{Eshky2018,
  author    = {Eshky, Aciel and Ribeiro, Manuel Sam and Cleland, Joanne and others},
  title     = {{UltraSuite}: A Repository of Ultrasound and Acoustic Data from Child Speech Therapy Sessions},
  booktitle = {Proceedings of Interspeech 2018},
  year      = {2018},
  pages     = {1888--1892},
  doi       = {10.21437/Interspeech.2018-1736}
}

@Article{Cleland2019,
  author  = {Cleland, Joanne and Scobbie, James M. and Roxburgh, Zoe and others},
  title   = {Enabling New Articulatory Gestures in Children With Persistent Speech Sound Disorders Using Ultrasound Visual Biofeedback},
  journal = {Journal of Speech, Language, and Hearing Research},
  volume  = {62},
  number  = {2},
  pages   = {229--246},
  year    = {2019},
  doi     = {10.1044/2018_JSLHR-S-17-0360}
}

@Article{Cleland2021,
  author  = {Cleland, Joanne and Scobbie, James M.},
  title   = {The Dorsal Differentiation of Velar From Alveolar Stops in Typically Developing Children and Children With Persistent Velar Fronting},
  journal = {Journal of Speech, Language, and Hearing Research},
  volume  = {64},
  number  = {6S},
  pages   = {2347--2362},
  year    = {2021},
  doi     = {10.1044/2020_JSLHR-20-00373}
}

@inproceedings{Hernandez2026PhonoQ,
  title     = {{Multilingual Phonological Feature Recognition with Self-Supervised Speech Models}},
  author    = {Abner Hernandez and Tomás Arias-Vergara and Daiqi Liu and Andreas Maier and Paula Andrea Pérez-Toro},
  year      = {2026},
  booktitle = {{Interspeech 2026}},
  pages     = {4014--4018},
  doi       = {10.21437/Interspeech.2026-2735},
  issn      = {2958-1796},
}

@InProceedings{Zheng2026CHILDESAligned,
  author       = {Zheng, Haolong and Hu, Yuanzhuo and Liang, Xinyu and others},
  title        = {{CHILDES-Aligned}: A Curated Children's Speech Dataset via Multi-Model Timestamp Ensembling},
  booktitle    = {Proc. IEEE Spoken Language Technology Workshop (SLT)},
  organization = {IEEE},
  year         = {2026},
  note         = {to appear; arXiv:2607.03670}
}

@Book{arias2022analysis,
  author    = {Arias-Vergara, Tom{\'a}s},
  title     = {Analysis of Pathological Speech Signals},
  series    = {Studien zur Mustererkennung},
  publisher = {Logos Verlag Berlin},
  address   = {Berlin, Germany},
  year      = {2022},
  isbn      = {978-3-8325-5561-0}
}

@inproceedings{arias2023measuring,
  title     = {{Measuring Phonological Precision in Children with Cleft Lip and Palate}},
  author    = {Arias-Vergara, Tom{\'a}s and Londo{\~n}o-Mora, Elizabeth and P{\'e}rez-Toro, Paula Andrea and others},
  year      = {2023},
  booktitle = {{Interspeech 2023}},
  pages     = {4638--4642},
  doi       = {10.21437/Interspeech.2023-2099},
  issn      = {2958-1796},
}

@InProceedings{Fan2024,
  author    = {Fan, Ruchao and Balaji Shankar, Natarajan and Alwan, Abeer},
  title     = {Benchmarking Children's ASR with Supervised and Self-Supervised Speech Foundation Models},
  booktitle = {Proceedings of Interspeech 2024},
  year      = {2024},
  pages     = {5173--5177},
  doi       = {10.21437/Interspeech.2024-1353}
}

@InProceedings{BlockMedin2024,
  author    = {Block Medin, Lucas and Pellegrini, Thomas and Gelin, Lucile},
  title     = {Self-Supervised Models for Phoneme Recognition: Applications in Children's Speech for Reading Learning},
  booktitle = {Proceedings of Interspeech 2024},
  year      = {2024},
  pages     = {5168--5172},
  doi       = {10.21437/Interspeech.2024-1095}
}

@InProceedings{Horii2025,
  author    = {Horii, Koharu and Tawara, Naohiro and Ogawa, Atsunori and others},
  title     = {Why is Children's ASR so Difficult? Analyzing Children's Phonological Error Patterns Using SSL-Based Phoneme Recognizers},
  booktitle = {Proceedings of Interspeech 2025},
  year      = {2025},
  pages     = {2870--2874},
  doi       = {10.21437/Interspeech.2025-2645}
}

@InProceedings{Getman2022,
  author    = {Getman, Yaroslav and Al-Ghezi, Ragheb and Voskoboinik, Ekaterina and others},
  title     = {wav2vec2-Based Speech Rating System for Children with Speech Sound Disorder},
  booktitle = {Proceedings of Interspeech 2022},
  year      = {2022},
  pages     = {3618--3622},
  doi       = {10.21437/Interspeech.2022-10103}
}

@InProceedings{Ng2022,
  author    = {Ng, Si-Ioi and Ng, Cymie Wing-Yee and Wang, Jiarui and others},
  title     = {Automatic Detection of Speech Sound Disorder in Child Speech Using Posterior-Based Speaker Representations},
  booktitle = {Proceedings of Interspeech 2022},
  year      = {2022},
  pages     = {2853--2857},
  doi       = {10.21437/Interspeech.2022-935}
}

@InProceedings{Sung2025,
  author    = {Sung, Selina S. and Ha, Seunghee and Yoon, Tae-Jin and others},
  title     = {Multitask Learning with Fused Attention for Improved ASR and Mispronunciation Detection in Children's Speech Sound Disorders},
  booktitle = {Proceedings of Interspeech 2025},
  year      = {2025},
  pages     = {5698--5702},
  doi       = {10.21437/Interspeech.2025-259}
}

@InProceedings{Choi2026,
  title     = {[b] = [d] - [t] + [p]: Self-supervised Speech Models Discover Phonological Vector Arithmetic},
  author    = {Choi, Kwanghee and Yeo, Eunjung and Cho, Cheol Jun and others},
  editor    = {Liakata, Maria and Moreira, Viviane P. and Zhang, Jiajun and others},
  booktitle = {Findings of the {A}ssociation for {C}omputational {L}inguistics: {ACL} 2026},
  month     = jul,
  year      = {2026},
  address   = {San Diego, California, United States},
  publisher = {Association for Computational Linguistics},
  url       = {https://aclanthology.org/2026.findings-acl.537/},
  doi       = {10.18653/v1/2026.findings-acl.537},
  pages     = {11048--11069},
  ISBN      = {979-8-89176-395-1}
}

\end{document}